\documentclass[graybox]{svmult}

\usepackage{type1cm}        
\usepackage{makeidx}         
\usepackage{graphicx}        
\usepackage{multicol}        
\usepackage[bottom]{footmisc}

\usepackage{newtxtext}       %
\usepackage[varvw]{newtxmath}       

\makeindex             

\usepackage{url}

\begin{document}

\title*{Towards Personalised Patient Risk Prediction Using
Temporal Hospital Data Trajectories}
\titlerunning{Towards Personalised Patient Risk Prediction Using
Temporal Data Trajectories}
\author{Thea Barnes, Enrico Werner, Jeffrey N. Clark, Raul Santos-Rodriguez}
\authorrunning{Thea Barnes, Enrico Werner, Jeffrey N. Clark, Raul Santos-Rodriguez} 
\institute{
Thea Barnes \at University of Bristol, Bristol (UK),
\newline\email{
dj19170@bristol.ac.uk} 
\and
Enrico Werner \at University of Bristol, Bristol (UK), \newline\email{enrico.werner@bristol.ac.uk}
\and Jeffrey N. Clark \at University of Bristol, Bristol (UK), \newline\email{jeff.clark@bristol.ac.uk} 
\and  Raul Santos-Rodriguez \at University of Bristol, Bristol (UK), \newline\email{enrsr@bristol.ac.uk}}
 
%
%
\maketitle


\abstract{Quantifying a patient's health status provides clinicians with insight into patient risk, and the ability to better triage and manage resources. Early Warning Scores (EWS) are widely deployed to measure overall health status, and risk of adverse outcomes, in hospital patients. However, current EWS are limited both by their lack of personalisation and use of static observations. We propose a pipeline that groups intensive care unit patients by the trajectories of observations data throughout their stay as a basis for the development of personalised risk predictions. Feature importance is considered to provide model explainability. Using the MIMIC-IV dataset, six clusters were identified, capturing differences in disease codes, observations, lengths of admissions and outcomes.
Applying the pipeline to data from just the first four hours of each ICU stay assigns the majority of patients to the same cluster as when the entire stay duration is considered. In-hospital mortality prediction models trained on individual clusters had higher F1 score performance in five of the six clusters when compared against the unclustered patient cohort. The pipeline could form the basis of a clinical decision support tool, working to improve the clinical characterisation of risk groups and the early detection of patient deterioration.}

\keywords{Clustering, Clinical Evaluation,  Explainability, Patient Subtypes}
\section{Introduction}

In recent years, hospitals have transitioned from paper-based to electronic health records (EHR), enabling automatic data-processing to support clinical decisions~\cite{vranas2017identifying, mcwilliams2019towards, castela2021identifying, carr2021evaluation, OEI2021100042}. To assist clinical decision-making, several Early Warning Scores (EWS) have been developed and deployed, including the National Early Warning Score 2 (NEWS2) which supports clinicians in the UK~\cite{balkan2018evaluating, subbe2001validation, rothman2013development, khwannimit2007comparison}. These scoring systems make use of routinely collected data to measure a patient’s overall health, with the aim of allowing early detection of patient deterioration towards in-hospital mortality, admission to ICU and/or cardiac arrest~\cite{gerry2020early}.

Whilst the simplicity of the metric makes it easily understood, interpreted, and trusted by clinicians, the lack of patient personalisation means that the scoring system is less effective for certain patient subgroups such as the elderly, pregnant, paediatric, and COVID-19-positive patients~\cite{Downey2017,carr2021evaluation}. Furthermore, EWS are limited through their evaluation of instantaneous measurements~\cite{YoussefAliAmer2020}. We argue that ignoring the temporal dimension is suboptimal by not utilising similarities in patient trajectories observed in medical environments.

Previous studies have utilised patients' medical history and temporal EHR data, such as to predict hospital admissions~\cite{arandjelovic2015discovering}, detect cardiovascular events and survival analysis~\cite{lee2019dynamic}.
Clustering algorithms have been applied to identify groups of patients such as those in ICU~\cite{vranas2017identifying}, ward patients~\cite{werner2023identification}, and cardiovascular clusters in sepsis patients~\cite{geri2019cardiovascular}, yet most studies have focused on static sets of observations. Few studies have applied clustering with longitudinal trajectories~\cite{aguiar2020phenotyping, rocheteau2023dynamic}. Accounting for and embedding temporal information into patient subtyping could provide insights into the progression of deterioration, allowing earlier detection of serious adverse events in patients.

Hence, this study presents a pipeline in which patients are clustered based on the trajectories of their vital signs over the course of their stay in ICU. The identified clusters are then applied for prediction of in-hospital mortality risk, a frequent component of early warning scores. Explainability techniques are embedded to aid clinicians in understanding the models. With further development the presented pipeline could become a decision support tool by providing a personalised EWS, improving patient care.

\begin{table*}[!ht]
\def\arraystretch{1.7}
\centering
\begin{tabular}{p{19mm}|p{15mm}|p{15mm}|p{15mm}|p{15mm}|p{15mm}|p{15mm}}
\hline
 &Cluster 0 &Cluster 1 &Cluster 2 &Cluster 3 & Cluster 4 & Cluster 5\\
 \hline
No. patients & \textbf{3955} & 308 & 284 & 620 & 345 & 250\\
Mortality ($\%$) & 6.3 & 26.0 & \textbf{58.0} & 4.8 & 7.8 & 18.0 \\
Age & 63.7 ($\pm 17.0$) & 61.8 ($\pm 15.4$) & \textbf{68.6} ($\pm 14.5$) &62.3 ($\pm 16.5$) & 59.2 ($\pm 16.8$) & 63.1 ($\pm 16.9$)\\
Leading \newline diagnosis & Pleural \newline effusion & AVD & Arrhythmia & Heart \newline disease & IBD & Lupus \\
No. diagnosis codes & 17.1 \newline ($\pm 8.7$) & 23.5 \newline ($\pm 9.0$) & 18.4 \newline ($\pm 9.0$) & 18.5 \newline ($\pm 8.4$) &19.2 \newline ($\pm 9.1$) & \textbf{23.5} \newline ($\pm 9.3$) \\
Length of stay (days) & 2.3 \newline ($\pm 2.2$) & \textbf{11.1} \newline ($\pm 10.3$) & 5.3 \newline ($\pm 8.8$) & 4.9 \newline ($\pm 5.3$)& 4.4 \newline ($\pm 4.9$)& 9.0 \newline ($\pm 7.2$)\\
Temperature ($^{\circ}C$) & 36.8 \newline ($\pm 0.3$)& 37.2 \newline ($\pm 0.5$)  &\textbf{37.3} \newline ($\pm 0.6$)  &37.0 \newline ($\pm 0.4$) & 37.1 \newline ($\pm 0.4$)& 37.0 \newline ($\pm 0.5$)\\
Respiration (/min) & \textbf{17.0} \newline ($\pm 4.6$)& 14.1 \newline ($\pm 3.2$)& 16.7 \newline ($\pm 4.6$) &15.1 \newline ($\pm 3.5$)& 15.7 \newline ($\pm 3.7$)& 14.3 \newline ($\pm 3.0$)\\
Heart Rate (bpm) & 84.3 \newline ($\pm 19.9$)& \textbf{93.8} \newline ($\pm 19.5$)& 90.2 \newline ($\pm 21.3$)& 85.7 \newline ($\pm 20.6$) & 88.1 \newline ($\pm 20.5$) & 91.5 \newline ($\pm 21.1$)\\
SATS ($\%$) &95.9 \newline ($\pm 3.8$) & \textbf{97.9} \newline ($\pm 5.1$) &97.6 \newline ($\pm 6.7$) &97.5 \newline ($\pm 3.8$) & 97.8 \newline ($\pm 3.4$) & 97.5 \newline ($\pm 5.6$) \\
Systolic BP (mmHg)& \textbf{116} \newline ($\pm 17.7$) &111 \newline ($\pm 13.5$)&115 \newline ($\pm 18.1$) &115 \newline ($\pm 15.0$)& 114 \newline ($\pm 15.1$)& 110\newline ($\pm 12.5$)  \\
GCS verbal & 4.7 \newline ($\pm 0.8$) & 2.2 \newline ($\pm 1.5$) & \textbf{1.4} \newline ($\pm 1.0$)& 2.7 \newline ($\pm 1.7$) & 1.9 \newline ($\pm 1.5$) & 2.6 \newline ($\pm 1.7$)\\
GCS motor & 5.9 \newline ($\pm 0.3$) &4.0 \newline ($\pm 1.6$) & \textbf{3.2} \newline ($\pm 1.5$)& 4.9 \newline ($\pm 1.6$) & 5.4 \newline ($\pm 1.1$) & 4.9 \newline ($\pm 1.4$)\\
GCS eye & 3.9 \newline ($\pm 0.3$)& 2.0 \newline ($\pm 1.3$) & \textbf{1.1} \newline ($\pm 0.4$) & 3.0 \newline ($\pm 1.3$) & 3.8 \newline ($\pm 0.5$) & 3.2 \newline ($\pm 1.2$)\\
\end{tabular}
\caption{Clinical characterisation of clusters utilising entire intensive care unit stays. Measurements are given as: mean (standard deviation). Note that these values are presented to illustrate averages per cluster, clustering itself was carried out on the series of temporal features across each patient's entire stay, as described in Section~\ref{clustering_method}. `No. diagnosis codes' refers to the average number of International Classification of Diseases (ICD) codes per patient. The Glasgow coma scale (GCS) assesses coma severity based on eye, verbal, and motor criterion, where the lower the score, the lower the patient's conscious state. AVD = Aortic valve disorder, IBD = Inflammatory bowel disease.
}
\label{table: average cluster}

\end{table*}

\section{Methodology}
\subsection{Data Source}
Electronic health records (EHR) were extracted from the Medical Information Mart for Intensive Care (MIMIC)-IV 2.0, covering hospital admission to the intensive care unit (ICU) between 2008 and 2019. In order to reduce computation time, 6,000 patients were randomly selected for which all vital sign measurements were available: comprising 48,994 sets of vital measurement readings in total. Vital sign measurements are respiration rate, oxygen saturation (SATS), temperature, systolic blood pressure, heart rate, and the Glasgow Coma Scale (GCS) which consists of eye-opening, motor, and verbal responsiveness.

Though the vitals are collected routinely, individual vital measurements are usually not taken at the same time and sets of measurements are not collected at regular intervals. Measurements taken within a period of 30 mins were grouped together. Missing vitals were imputed by taking the mean value of the patient's stay. Following existing protocol~\cite{werner2023identification, werner2023explainable} features found to be normally distributed  (temperature, systolic blood pressure, respiration rate, and GCS eye) were scaled to unit variance, all others were scaled using the Min-Max scaler.
The dataset contained a mixture of ICD-9 and ICD-10 codes. ICD codes (International Classification of Diseases) are a global standard for diagnostic health information and have a tree-like structure. For uniformity and comparability, this article used only the top-level classification layer and converted all ICD codes to the ICD-9 code standard. Conversion was achieved through backward mapping using GEM files (general equivalence mapping), following established protocol~\cite{cartwright2013icd}.

\subsection{Clustering of Trajectories}
\label{clustering_method}
\begin{figure}[!ht]
\centering
\includegraphics[width=1\linewidth]{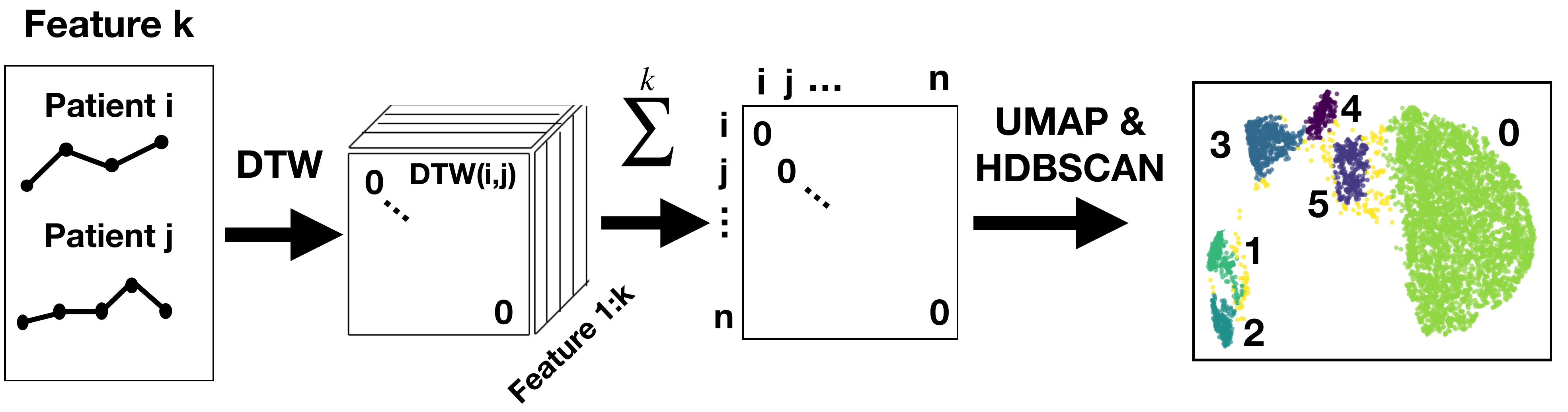}
\caption{Procedure for clustering of patients based on similarities of their vital sign trajectories. For each feature, each patient's time-series is compared using dynamic time warping (DTW). DTW(i,j) represents the Dynamic Time Warping distance between trajectories of feature $k$ for patients $i$ and $j$. The DTW matrix for each feature is summed together, resulting in a matrix for the total DTW distance across all features between all patients. Dimensionality reduction (UMAP) and clustering (HDBSCAN) are then computed. grouping patients on the similarity of their feature trajectories.}
\label{fig: pipeline overview}
\end{figure}

Clustering was carried out on the trajectories of temporally changing measurements.
In order to compute pairwise similarities between misaligned patient vital time series, independent Dynamic Time Warping (DTW) was used (Figure~\ref{fig: pipeline overview}). DTW finds the optimal alignment between two time-series by minimizing the distance/cost function. This technique allows the matching of patients with similar vital measurement progressions, albeit at different time intervals. DTW was computed independently for each vital sign and an overall distance matrix was computed by summing over the distance matrices for all vitals. This resulted in a 6,000 x 6,000 distance matrix of pairwise similarities between patient vitals. 

Uniform Manifold Approximation and Projection (UMAP) was used for dimension reduction as it preserves global and local structures~\cite{mcinnes2018umap}. Next, Hierarchical density-based spatial clustering of applications with noise (HDBSCAN*) was used to cluster the UMAP embeddings. UMAP coupled with HDBSCAN* is found to be the best pairing for time series clustering~\cite{allaoui2020considerably, pealat2022improved, pealat2021improved}. The number of clusters was evaluated based on their intra-cluster similarity using the Silhouette index, Calinksi-Harabasz index and the Davies-Bouldin index. Cluster separation was evaluated based on demographics, diagnosis, vital measurements and visualisation in the embedding space.

In addition to processing the entire duration of each patient's stay in ICU, we considered subsets of the stay duration, from admission to: 4 hours, 24 hours, 72 hours and 1 week since admission. This provided a means to assess how cluster assignment, and prediction quality, evolved as more of the patient stay is evaluated.

\subsection{Mortality Prediction}
After clustering, explainable boosting machine models were developed to predict risk of in-hospital mortality. In addition to the vital signs, features for risk prediction included patient characteristics: age and gender, admission details: first care-unit, last care-unit, admission type, admission location, length of stay, most occurring ICD code, second most occurring ICD code, and total number of unique ICD codes. Last care-unit and length of stay were not included in the models where reduced time frames of patient stays were considered. Feature importance was analysed for each cluster's model.
Class imbalance resulting from low mortality rates (Table \ref{table: average cluster}), was dealt with during training by implementing Synthetic Minority Over-sampling Technique for Nominal and Continuous (SMOTE-NC). Each model's performance was evaluated using stratified k-fold cross-validation with 10 folds. Predictive ability was compared between clusters against the entire unclustered patient cohort. We have made code available to run the entire pipeline
\footnote{\url{https://github.com/TheaRoseBarnes/Technical_Project_}}.

\begin{table*}[!ht]
\centering
\def\arraystretch{1.7}
\begin{tabular}{p{15mm}|p{13mm}|p{13mm}|p{13mm}|p{13mm}|p{13mm}|p{15mm}|p{13mm}|}
\hline
& Un-\newline clustered &Cluster 0 & Cluster 1 &Cluster 2 &Cluster 3 & Cluster 4 & Cluster 5\\
 \hline
AUROC & 0.945 (0.012) &\textbf{0.962} (0.009)  & 0.909 (0.056) & 0.914 (0.046) &0.909 (0.089)  & 0.923 (0.085)& 0.918 (0.059)\\
Accuracy  &0.926 (0.005) & 0.967 (0.006)& 0.885 (0.036) & 0.845 (0.059)  & \textbf{0.970} (0.018) & 0.953 (0.174)& 0.924 (0.028)\\
F1 & 0.824 (0.013) &0.883 (0.021)  &0.829 (0.055)  &0.844 (0.060) & 0.821 (0.113) &0.848 (0.054)& \textbf{0.864} (0.052)\\
Precision &0.627 (0.023) &0.749 (0.046) &0.751 (0.104)  &\textbf{0.862} (0.051) & 0.706 (0.209)  & 0.695 (0.130)& 0.765 (0.097) \\
Sensitivity & 0.768 (0.034)& 0.824 (0.037) &0.719 (0.100) & \textbf{0.843} (0.116)&0.634 (0.250) & 0.814 (0.158)& 0.800 (0.138)\\
Specificity &0.945 (0.005) &0.978 (0.005) &0.931 (0.104) & 0.939 (0.069) &\textbf{0.987} (0.009) & 0.967 (0.021)& 0.949 (0.026) \\
Brier score &0.055 (0.004) &0.026 (0.004) &0.097 (0.032) &0.133 (0.051) &\textbf{0.025} (0.016) & 0.039 (0.016)& 0.066 (0.023)\\
\end{tabular}
\caption{Explainable boosting machine model performance averaged over 10 train-test splits and provided as: mean (SD). The displayed clustering and predictions consider data across each patient's entire intensive care unit stay. Refer to the right-hand panel of Figure~\ref{fig: pipeline overview} for a visualisation of the cluster arrangement. 'Unclustered' refers to a model developed on the entire patient cohort.}
\label{mortaility prediction metrics}
\end{table*}

\begin{figure*}[!ht]
\centering
\includegraphics[width=1\linewidth]{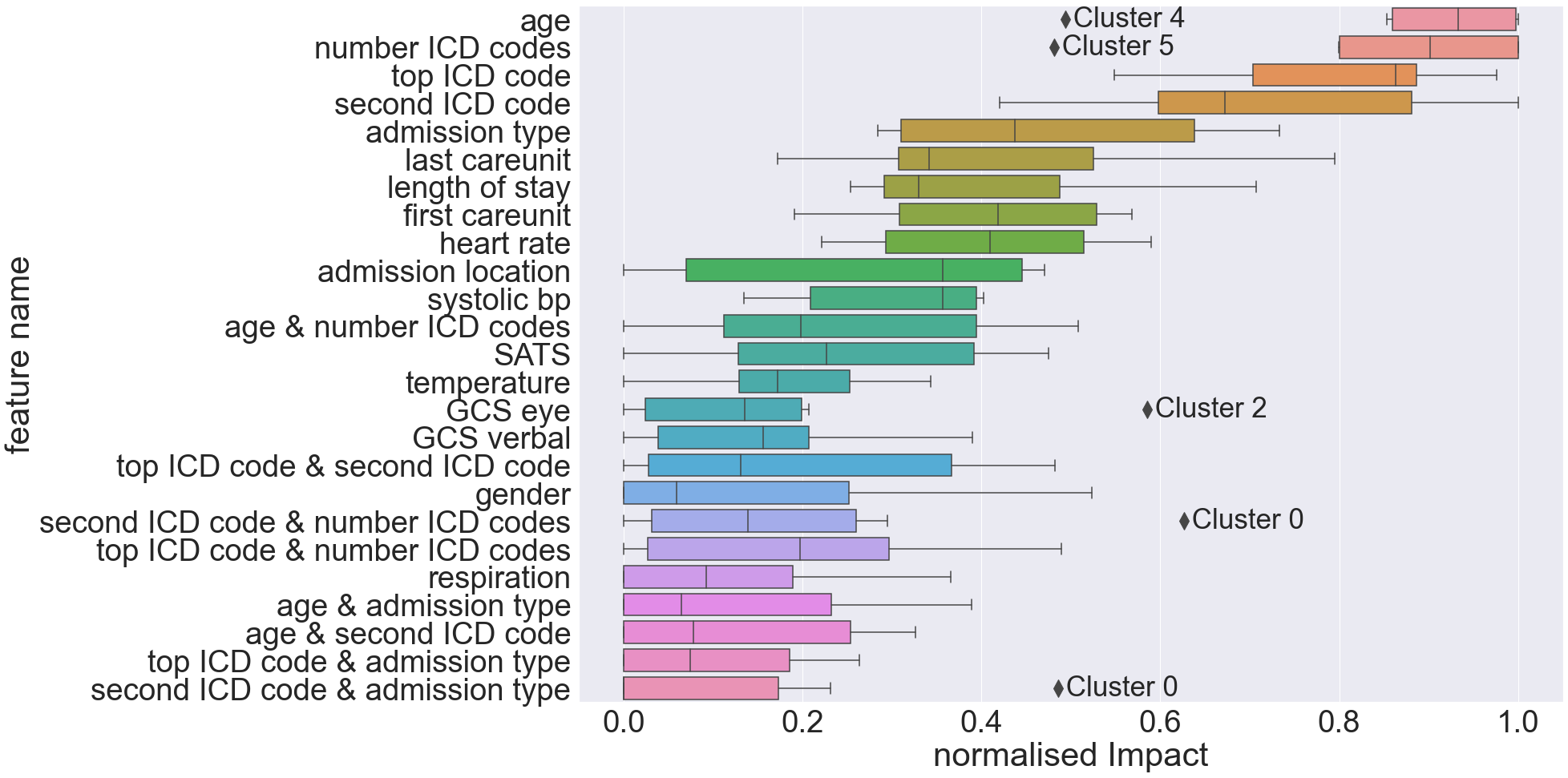}
\caption{Aggregated feature impact for in-hospital mortality prediction models when considering each patient's entire intensive care unit stay. Top and second ICD codes are the two most frequently occurring ICD codes throughout the entire ICU stay. bp = Blood pressure, SATS = Oxygen saturation, GCS = Glasgow Coma Scale.}
\label{fig: feature importances}
\end{figure*}

\section{Results and Discussion}
\subsection{Entire ICU Stay}
We identified six clusters with distinct characteristics in terms of vital signs, associated diagnostic codes, and length of stay and outcome (Right panel of Figure~\ref{fig: pipeline overview}, Table~\ref{table: average cluster}). Internal validity, and therefore the optimal number of clusters, was verified when the silhouette score was maximised, and the Calinksi-Harabasz index and Davis-Bouldin index reached a near local minimum at a minimum sample parameter of 60. The same number of clusters were identified in an independent study carried out across intensive care units at another US hospital system~\cite{vranas2017identifying}.

F1 scores computed for five of the six cluster classifier models suggest improved risk prediction performance compared to the classifier developed on the unclustered data (Table \ref{mortaility prediction metrics}). Models trained and tested on the data of patients belonging to cluster 2 averaged a precision of 0.862 up to a recall of 0.843 over ten train-test splits. Performance for at least one cluster appeared better than the unclustered data across all computed metrics. Using feature trajectories as part of the pipeline limits the ease of interpretability and explainability compared to using individual measurement readings to cluster and predict patient outcomes. Prior to deployment, further work should be carried out to understand and describe the patient subtypes identified by these clusters, as demonstrated by previous studies~\cite{vranas2017identifying, werner2023explainable}. Sub-clustering may enable the identification of more clearly defined sub-types~\cite{werner2023explainable}.

For most cluster models, age was the most important feature in predicting mortality, followed by the three features that include a reference to ICD codes which relate to patient diagnosis (Figure \ref{fig: feature importances}). Variations in the impact of features across the models are observed, demonstrated by the outliers where importance differs greatly between models. For instance, the two most important features for most clusters, age and the number of ICD codes, have an outlier each, suggesting these features are less important for clusters 4 and 5 respectively. Two of the interaction terms are more important in predictions for cluster 0 than models trained for other clusters. Future work could consider how some vital signs may be more important for certain patients than for others, for instance by computing their deviation from a healthy baseline and using this as a feature.
Missing values were imputed with the patient's average value across the entire ICU stay for that feature. This would be problematic for deployment and is therefore a limitation of the presented method. Future work should consider more appropriate methods of imputation, such using the last known value: forward fill.

Currently the pipeline has been implemented on a single dataset with relatively few patients for demonstration purposes. Future work should include analysis on additional datasets to explore generalisability of the presented results.


\begin{figure*}[!ht]
\centering
\includegraphics[width=\linewidth]{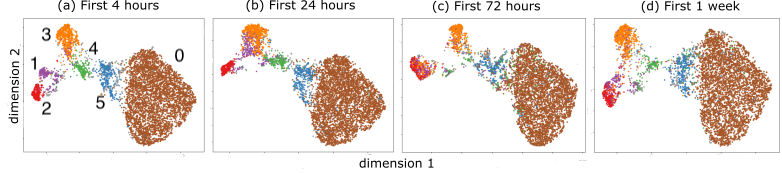}
\caption{Cluster evolution when considering data from varying lengths of time from first admission to ICU. Cluster labels in panel (a) correspond to cluster labels from the whole ICU stay (Figure~\ref{fig: pipeline overview}).}
\label{fig:cluster embeddings over time}
\end{figure*}


\subsection{Changing Time Horizons of ICU Stays}

Figure \ref{fig:cluster embeddings over time} visualises cluster evolution. Using data from a time frame of the first 4 hours since admission, the highest risk cluster as discovered with the entire ICU stay data is already distinct (cluster 2, red). Cluster 3, the lowest risk cluster, and cluster 1, the second highest risk cluster are initially distinct (Figure \ref{fig:cluster embeddings over time}a, top left) but merge over the 24 hour and 72 hour time frames. Once data comprising all measurements up to one week since admission are included, all six clusters have formed, with the majority of the same cluster labels as when the entire stay is used. However, of note clusters 1, 3, 4 and 5 appear least separated when using 24 and 72 hours of data (Fig~\ref{fig:cluster embeddings over time}b and c) compared to using less data (first four hours, Fig~\ref{fig:cluster embeddings over time}a) or more data (first week, Fig~\ref{fig:cluster embeddings over time}d). This suggests that the stages of patient journeys warrant investigation and could result in identifying the optimum length of time, or proportion of the total ICU stay, to use in the pipeline to results in the most distinct and clinically useful clusters. Future work should compare clustering and prediction performance when computed with all time points considered independent: i.e. not using trajectories. The time points of diagnoses and associated knowledge of the development of health issues during the ICU stay could also be added as features in an effort to better subtype and warn of patient deterioration. 

In general, it is observed that the embeddings of the reduced time frames of data all bear a strong resemblance to the structure and partitioning when utilising the data for the whole stay, despite the average length of stay for some clusters being several days longer than the trajectories considered in these reduced time frames (Table \ref{table: average cluster}). A patient from Cluster 0 is particularly likely to be consistently assigned at all time frames considered, with over 80\% of patients assigned to the same cluster at all time points. This suggests that certain patient subtypes can be successfully identified using a greatly limited subset of their trajectory data.

\begin{figure}[!ht]
\centering
\includegraphics[width=1\linewidth]{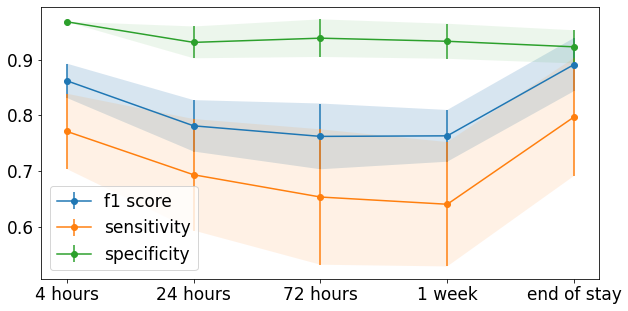}
\caption{Weighted average performance of in-hospital mortality models for varying time frames since admission. `end of stay' includes the entire patient stay and utilises additional features not available for the shorter time frames. Error bars are standard deviations.}
\label{fig:comparison}
\end{figure}

Average predictive performance varied across considered time frames (Figure \ref{fig:comparison}). When data from the entire stay for each patient is considered, the highest average sensitivity and F1 score were observed. Specificity remains high across all time frames and achieves the lowest standard deviation across clusters compared to the other performance metrics. When only the first four hours of ICU stay data is considered, the weighted average standard deviation of specificity is smallest. A trend is seen where the mean F1 scores and sensitivities are high for the first four hours since admission, drop for time frames of 24 hours, 72 hours and one week, and are optimal for the entire stay. 
It should be noted that the model for the entire stay includes additional features, such as length of stay, not available to the earlier time frames. However it is clear that good performance is achievable using smaller time frames of the patient stay, therefore providing evidence that such a pipeline may provide real-time utility to clinicians as a decision support tool.

\section{Conclusion}
We present a pipeline to subtype patients based on the trajectories of their routinely collected vital signs over the course of their stay in ICU, and to predict in-hospital mortality risk on these patient clusters. 

The pipeline is demonstrated in 48,994 sets of vital measurement readings across 6,000 patients' stays, revealing distinct patient characteristics and showing that subtyping patients prior to risk prediction can boost predictive performance.
Moreover, subsets of data from the entire stay are considered, down to the first four hours since admission, showing how patient trajectories and risk prediction evolve during their stay in ICU. 

Future work should validate these findings from a larger cohort of patients from at least one alternative hospital, alongside addressing limitations of the presented study and suggestions for future work as described. A clinical decision support tool utilising the temporal dimension and grouping patients on their common trajectories, as presented here, could contribute to improving the personalisation, and predictive performance of Early Warning Scores, resulting in improved patient risk prediction.

\section*{Acknowledgments}
JNC and RSR are funded by the UKRI Turing AI Fellowship [grant number EP/V024817/1].

\bibliographystyle{spmpsci}
\bibliography{main}

\end{document}